\documentclass{llncs}
\usepackage{graphics,graphicx}
\graphicspath{{./images/}}
\usepackage{amsmath}
\usepackage{wrapfig}
\usepackage{subcaption} 
\usepackage{xcolor}
\usepackage{siunitx}
\usepackage{subcaption}

\usepackage{amssymb,bm,amsmath,amsfonts,mathrsfs}
\usepackage[ruled, vlined]{algorithm2e}

\begin{document}

\title{Globally Optimal Segmentation of Mutually Interacting Surfaces using Deep Learning}

\author{
	Hui Xie\inst{5} \and
	Zhe Pan, MD \inst{1,2} \and 
	Leixin Zhou\inst{5}\and
	Fahim A Zaman\inst{5} \and
	Danny Chen \inst{4} \and
	Jost B Jonas, MD \inst{1,3}\and
	Yaxing Wang, MD \inst{1} \and
	Xiaodong Wu\inst{5}
}
\authorrunning{H. Xie, et al.}
\institute{
	Beijing Institute of Ophthalmology, Beijing Tongren Hospital, Capital University of Medical Science, Beijing Ophthalmology and Visual Sciences Key Laboratory, Beijing, China \and
	Eye Hospital of China Academy of Chinese Medical Sciences, Beijing, China \and
	Department of Ophthalmology, Medical Faculty Mannheim, Heidelberg University, Mannheim, Germany \and
	Department of Computer Science and Engineering, University of Notre Dame \and
	Department of ECE, University of Iowa, Iowa City, IA 52242, USA 
}

\maketitle
\begin{abstract}
Segmentation of multiple surfaces in medical images is a challenging problem, further complicated by the frequent presence of weak boundary and mutual influence between adjacent objects. The traditional graph-based optimal surface segmentation method has proven its effectiveness with its ability of capturing various surface priors in a uniform graph model. However, its efficacy heavily relies on handcrafted features that are used to define the surface cost for the  ``goodness'' of a surface. Recently, deep learning (DL) is emerging as powerful tools for medical image segmentation thanks to its superior feature learning capability. Unfortunately, due to the scarcity of training data in medical imaging, it is nontrivial for DL networks to {\em implicitly} learn the global structure of the target surfaces, including surface interactions. In this work, we propose to parameterize the surface cost functions in the graph model and leverage DL to learn those parameters. The multiple optimal surfaces are then simultaneously detected by minimizing the total surface cost while {explicitly} enforcing the mutual surface interaction constraints.  The optimization problem is solved by the primal-dual Internal Point Method, which can be implemented by a layer of neural networks, enabling efficient end-to-end training of the whole network. Experiments on Spectral Domain Optical Coherence Tomography (SD-OCT) retinal layer segmentation and Intravascular Ultrasound (IVUS) vessel wall segmentation demonstrated very promising results. All source code is public at \cite{MoDL-OSSeg} to facilitate further research at this direction. 


\end{abstract}

\section{Introduction}
The task of optimally delineating 3D surfaces representing object boundaries is important in segmentation and quantitative analysis of volumetric medical images. OCT image segmentation to detect and localize the intra-retinal boundaries is a necessary basis for ophthalmologists in diagnosis and treatment of retinal pathologies \cite{KafiehOCTAlgReview2013}, e.g glaucoma. IVUS image segmentation produces cross-sectional images of blood vessels that provide quantitative assessment of the vascular wall, information about the nature of atherosclerotic lesions as well as plaque shape and size \cite{CardinalIVUSSegmentation2006}. Interactive surfaces segmentation technologies are applied tissues layers segmentation in medical images.

In medical imaging, many surfaces that need to be identified appear in mutual interactions. These surfaces are ``coupled'' in a way that their topology and relative positions are usually known already (at least in a general sense),
and the distances between them are within some specific range. Clearly, incorporating these surface-interrelations into the segmentation can further improve its accuracy and robustness, especially when insufficient image-derived information is available for defining some object boundaries or surfaces.  Such insufficiency can be remedied by using clues from other related boundaries or surfaces. Simultaneous optimal detection of multiple coupled surfaces thus yields superior results compared to the traditional single-surface detection approaches. Simultaneous segmentation of coupled surfaces in volumetric medical images is an under-explored topic, especially when more than two surfaces are involved.

Several approaches for detecting coupled surfaces have been proposed in past years. The graph-based methods \cite{li2006optimal,song2012optimal,shah2019optimal} have been proven one of the state-of-the-art {\em traditional} approaches for surface segmentation in medical images. The great success of the methods is mainly due to their capability of modeling the boundary surfaces of multiple interacting objects, as well as {\em a prior} knowledge reflecting anatomic information in a complex multi-layered graph model, enabling the segmentation of all desired surfaces to be performed simultaneously in a single optimization process with guaranteed global optimality. The essence of the graph model is to encode the surface cost, which  measures the ``goodness'' of a feasible surface based on a set of derived image features, as well as the surface interacting constraints, into a graph structure.  The major drawback is associated with the need for handcrafted features to define the surface cost of the underlying graphical model.

Armed with superior data representation learning capability, deep learning (DL) methods are emerging as powerful alternatives to traditional segmentation algorithms for many medical image segmentation tasks~\cite{litjens2017survey,shen2017deep}. The state-of-the-art DL segmentation methods in medical imaging include fully convolutional networks (FCNs)~\cite{long2015fully} and U-net~\cite{ronneberger2015u}, which model the segmentation problem as a pixel-wise or voxel-wise classification problem. However, due to the scarcity of training data in medical imaging, it is at least nontrivial for the convolutional neural networks (CNNs) to {\em implicitly} learn the global structures of the target objects, such as shape, boundary smoothness and interaction. Shah \textit{et al}. \cite{shah2018multiple} first formulated the single surface segmentation as a regression problem using an FCN followed by fully-connected layers to enforce the monotonicity of the target surface.  More recently, Yufan He \textit{et al.}~\cite{YufanHe2019} utilized a U-net as a backbone network to model the multiple surface segmentation with regression by a fully differentiable soft argmax, in which the ordering of those surfaces is adjusted to be guaranteed by a sequence of ReLU operations. 

In this work, we propose to unify the powerful feature learning capability of DL with the  successful  graph-based surface segmentation model in a single deep neural network for end-to-end training to achieve globally optimal segmentation of multiple interacting surfaces. In the proposed segmentation framework, the surface costs are parameterized and the DL network is leveraged to learn the model from the training data to determine the parameters for the input image. The multi-surface inference by minimizing the total surface cost while satisfying the surface interacting constraints is realized by the primal-dual Internal Point Method (IPM) for constrained convex optimization, which can be implemented by a layer of neural networks enabling efficient back-propagation of gradients with virtually no additional cost~\cite{OptNet2017}. Thus, the DL network for surface cost parameterization can be seamlessly integrated with the multi-surface inference to achieve the end-to-end training.

\section{Methods}
To clearly present the essence of the proposed surface segmentation framework, we consider the simultaneous segmentation of multiple {\em terrain-like} surfaces. For the objects with complex shapes, the unfolding techniques~\cite{zhou20193} developed for the graph-based surface segmentation methods as well as the convolution-friendly resampling approach~\cite{yin2010logismos,oguz2014logismos}, can be applied. 

\subsection{Problem Formulation}
Let $\mathcal{I}(X, Y, Z)$ of size $X$$\times$$Y$$\times$$Z$ be a given 3-D volumetric image. For each $(x, y)$ pair, the voxel subset $\{\mathcal{I}(x, y, z) | 0 \leq z < Z\}$ forms a column parallel to the $\mathbf{z}$-axis, denoted by $q(x, y)$, which is relaxed as a line segment from $\mathcal{I}(x, y, 0)$ to $\mathcal{I}(x, y, Z-1)$. Our target is to find $N>1$ terrain-like surfaces $\vec{S} =\{S_0, S_1, \ldots, S_{N-1}\}$, each of which intersects every column $q(x, y)$ at exactly one point. %(Fig.~\ref{fig:surf_def}). 

In the graph-based surface segmentation model~\cite{li2006optimal,song2012optimal,shah2019optimal}, each voxel $\mathcal{I}(x,y,z)$ is associated with an on-surface cost $c_i(x,y,z)$ for each sought surface $S_i$, which is inversely related to the likelihood that the desired surface $S_i$ contains the voxel, and is computed based on handcrafted image features. The {\em surface cost} of $S_i$ is the total on-surface cost of all voxels on $S_i$. The on-surface cost function $c_i(x,y,z)$ for the column $q(x, y)$ can be an arbitrary function in the graph model. However, an ideal cost function $c_i(x,y,z)$ should express a type of convexity: as we aim to minimize the surface cost, $c_i(x,y,z)$ should be low at the surface location; while the distance increases from the surface location along the column $q(x, y)$, the cost should increase proportionally. We propose to leverage DL networks to learn a Gaussian distribution $\mathcal{G}(\mu_i(q), \sigma_i(q))$ to model the on-surface cost function $c_i(x,y,z)$ for each column $q(x, y)$, that is, $c_i(x,y,z) = \frac{(s_i-\mu_i)^2}{2\sigma_i^2}$. Thus, the surface cost of $S_i$ is parameterized with $(\vec{\mu}_i, \vec{\sigma}_i)$. 

For multiple surfaces segmentation, a surface interacting constraint is added to every column $q(x, y)$ for each pair of the sought surfaces $S_i$ and $S_j$. For each $q(x, y)$, we have $\delta_{ij}(q) \le S_i(q) - S_j(q) \le \Delta_{ij}(q)$, where $\delta_{ij}(q)$ and $\Delta_{ij}(q)$ are two specified minimum and maximum distance between $S_i$ and $S_j$, respectively, with $S_i$ on top of $S_j$. The multi-surface segmentation is formulated as an optimization problem, where the parameterized surface costs are derived using deep CNNs:

\begin{equation} \label{eq:model}
\begin{aligned}
\vec{S^*} = &\operatorname*{argmin}_{\vec{S}} \sum_{i=0}^{N-1}\sum_{\mathcal{I}(x, y, z) \in S_i} c_i(x, y, z)_{|(\vec{\mu}_i, \vec{\sigma}_i)}  \\
&s.t. \quad \delta_{ij}(q) \le S_i(q) - S_j(q) \le \Delta_{ij}(q) \quad \forall i, j, q
\end{aligned}
\end{equation}

\subsection{The Surface Segmentation Network Architecture}
\begin{figure}
	\includegraphics[width=\textwidth]{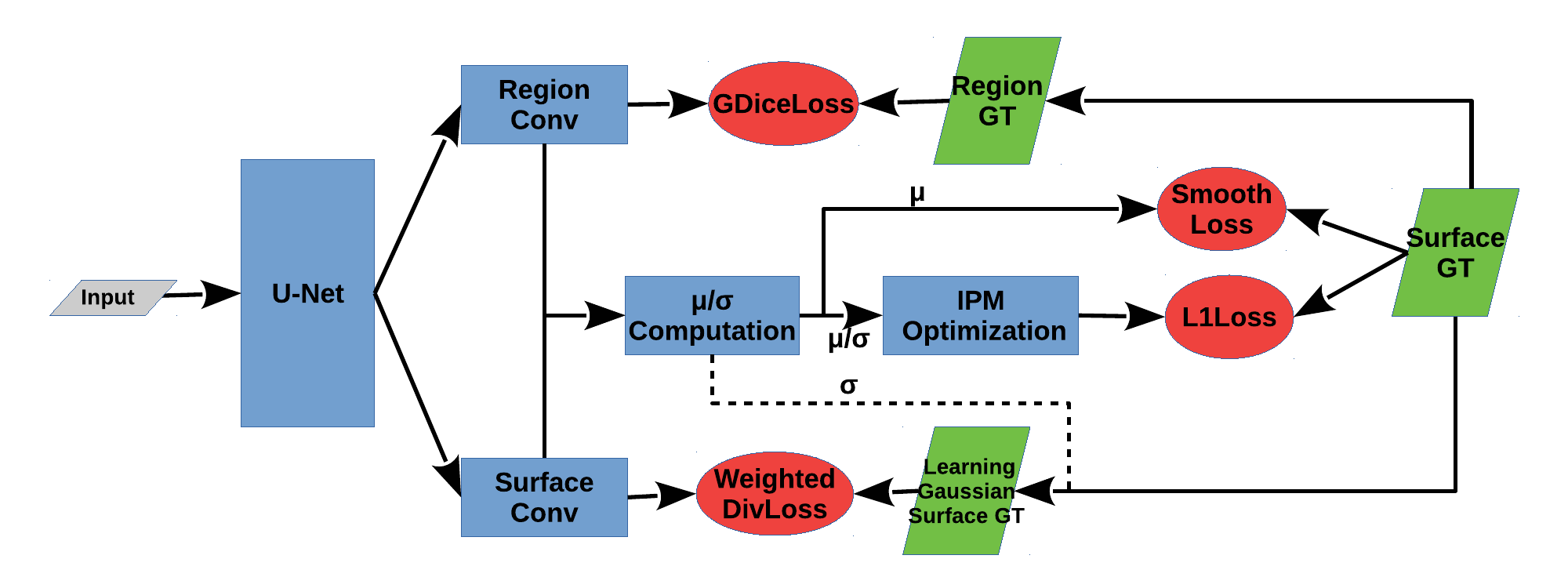}
	\caption{Illustration of the network architecture of the proposed multiple surface segmentation.}
	\label{SystemArch}
\end{figure}

As shown in Fig.~\ref{SystemArch}, our segmentation network consists of two integrative components: One aim to learn the surface cost parameterization $(\vec{\mu}, \vec{\sigma})$ in Eqn.~(\ref{eq:model}); the other strikes to solve the the optimal surface interference by optimizing Eqn.~(\ref{eq:model}) with the IPM optimization module. The surface cost is parameterized with ($\vec{\mu}$, $\vec{\sigma}$), which models the Gaussian distribution of the surface locations along each image column. The assumption behind this learning Gaussian surface ground truth is that the predicted surface locations in the H dimension should has the maximum probability, while locations deviating from this predicted surface location have smaller, instead of zero, probability as the predicted result. Bigger variance means most difficult recognizing surface, while small variance means easy recognizing surface. RegionConv is a convolution module to output $(N+1)$-region segmentation, while SurfaceConv is a convolution module to output $N$-surface segmentation probability distribution. IPM Optimization indicates primal-dual Internal Point Method for constrained convex optimization. Input includes raw image, gradient of a raw image along H, W dimension, and magnitude and direction of the gradient, total 5 channels. GDiceLoss is an $(N+1)$-class Generalized Dice Loss. Weighed DivLoss is an image-gradient weighted divergence loss. GT denotes ground truth. Dashed line indicates optional for different experiments. Thus, the whole network can then be trained in an end-to-end fashion and outputs globally optimal solutions for the multiple surface segmentation.

\noindent
{\bf Surface Cost Parameterization.} We utilize U-net~\cite{ronneberger2015u} as the backbone of our deep network for the feature extraction. The implemented U-net has seven layers with long skip connections between the corresponding blocks of its encoder and decoder. Each block has three convolution layers with a residual connection~\cite{Residual2016}. The output feature maps of the U-net module is then fed into the following RegionConv and SurfaceConv modules (Fig.~\ref{SystemArch}). The RegionConv module is implemented with three-layer convolutions followed by a $1$$\times$$1$ convolution and softmax to obtain the probability maps for the $(N+1)$ regions divided by the sought $N$ surfaces. The SurfaceConv module is implemented with the same module structure of RegionConv to compute the location probability distribution along every image column for each surface. Note that each sought surface intersects every image column exactly once.

The RegionConv module directly makes use of the region information, which may help direct the U-net learning robust features for surface segmentation. In addition, the output region segmentation is used to estimate the surface locations. For each sought surface $S_i$ and every image column $q$, the estimated surface location $\gamma_i$ is the average envelop of the $(i+1)$-th region on column $q$, as there is no guarantee that each of the  predicted $(\lambda+1)$ regions is consecutive along the column based on voxel-wise classification by RegionConv, especially in the early stage of the training process. We also calculate a confidence index $c$ ($0 \le c \le 1$) for the surface location estimation $\gamma_i$ based on the number of region disordering with $c = 1$ for no disordering. 

For each surface $S_i$, based on the surface location probability $p_i(z)$  on every image column $q(x, y)$ from the SurfaceConv module, the expected surface location $\xi_i = \sum_{z = 0}^{Z-1} z * p_i(z)$. Combined with the RegionConv module, the surface location distribution of $S_i$ on column $q$ is modeled with a Gaussian $\mathcal{G}_i(\mu_i, \sigma_i)$, as follows.
\begin{equation} \label{formula:mu} 
\mu_i = \frac{c\gamma_i + (\kappa - c)\xi_i}{\kappa},
\end{equation}
\begin{equation} \label{formula:sigma} 
\sigma_i^2 = \sum_{z=0}^{Z-1} p_i(z)*(z-\mu_i)^2,
\end{equation}
where $\kappa \ge 2$ is used to balance the fidelity of information from RegionConv and SurfaceConv. Thus, the surface cost $\sum_{\mathcal{I}(x, y, z) \in S_i} c_i(x, y, z)_{|(\vec{\mu}_i, \vec{\sigma}_i)}$ of surface $S_i$ is parameterized with $(\vec{\mu}_i, \vec{\sigma}_i)$.

\noindent
{\bf Globally Optimal Multiple Surface Inference.} Given the surface cost parameterization $(\vec{\mu}, \vec{\sigma})$, the inference of optimal multiple surfaces can be solved by optimizing Eqn.~(\ref{eq:model}), which is a constrained convex optimization problem. In order to achieve an end-to-end training, the optimization inference needs to be able to provide gradient back-propagation, which impedes the use of traditional convex optimization techniques. We exploit the OptNet technique~\cite{OptNet2017} to integrate a primal-dual interior point method (IPM) for solving Eqn.~(\ref{eq:model}) as an individual layer in our surface segmentation network (Fig.~\ref{SystemArch}). Based on Amos and Kolter's theorem~\cite{OptNet2017}, the residual equation $\vec{r}(\vec{z}, \vec{\theta})$ to Eqn.~(\ref{eq:model}) derived from the Karush-Kuhn-Tucker conditions at the optimal solution $\vec{z}^*$ can be converted into a full differential equation $\bm{J} \left[ \begin{smallmatrix} \vec{dz}\\ \vec{d\vec{\theta}} \end{smallmatrix} \right]= \vec{0}$, where $\bm{J}$ is a Jacobian of $\vec{r(z,\theta)}$ with respect to $\vec{(z,\theta)}$, $\vec{\theta}$ is the input to the IPM optimization module including $(\vec{\mu}, \vec{\sigma})$, and $\vec{z}$ defines the surface locations of all $\lambda$ surfaces. We thus can deduce partial differentials $\frac{\vec{dz}}{\vec{d\theta}}$, which can be used to compute the back-propagation gradients $\frac{dL}{\vec{d\theta}} = \frac{dL}{\vec{dz}} \frac{\vec{dz}}{\vec{d\theta}}$, where $L$ is the training loss. IPM method is a 2nd order Newton method with complicated matrix inversion, but which just needs less than 10 iterations to get converge in our context, so it still supports high-epoch training form. In test,it just needs about 1 minute for each volume image. Please refer to the  Appendix for a detailed IPM algorithm,  and all codes including preprocessing, network, optimization, and config file are publice at \cite{MoDL-OSSeg}. 

\subsection{Network Training Strategy}
Multiple loss functions are introduced to focus on the training of different modules in the proposed multiple surface segmentation network (Fig.~\ref{SystemArch}). In the proposed SurfaceConv module, the softmax layer works on each image {\em column}, not on each voxel. The rational is that we assume each target surface intersects with each column by exactly once, and so the probabilities are normalized within each column. We assume SurfaceConv should output a Gaussian shaped probability map for each column, which mimics the Bayesian learning  for each column and shares merits with knowledge distillation~\cite{hinton15} and distillation defense~\cite{papernot2016distillation}.

To encourage SurfaceConv outputs reasonable probability maps, an innovative weighted divergence loss $L_{Div}$ is utilized for SurfaceConv training. It inherits from KLDLoss (Kullback–Leibler divergence). It also measures distribution distance between 2 distribution, but it more emphasizes probability consistence of some weighed critical points between 2 distributions. 
$L_{Div} = \sum_i w_ig_i \|log(\frac{g_i}{p_i})\|,$ where $i$ indicates all pixels in $N$ classes, and $g_{i}$ is ground truth probability at pixel $i$, $p_{i}$ is predicted probability at pixel $i$, $w_i \in W$ is a pixel-wise weight from raw image gradient magnitude: $W = 1 + \alpha\|\nabla(I)\|$,where $\alpha =10$ as an experience parameter. In our applications, we hope the better probability consistence at pixels of bigger image gradients between the prediction and ground truth. We use the surface location of each reference surface on each column as $\mu$ and use either fixed $\sigma$ or dynamically from the the $\vec{\mu}$/$\vec{\sigma}$ computation module to form the ground truth Gaussian distribution. 

For the RegionConv module, a generalized Dice loss $L_{GDice}$ \cite{GDL2017} is introduced to counter the possible high unbalance in region sizes.

For the predicted surface locations, in addition to using $L_1$-loss $L_1$ to measure the difference between the prediction and the surface ground truth, we introduce a novel SmoothLoss $L_{smooth}$ to regularize the smoothness and mutual interaction of sought surfaces. More precisely, $L_{smooth}$ is the total sum of the mean-squared-errors (MSEs) of the surface location changes between any two adjacent image columns to the ground truth, plus the total sum of the MSEs of thickness on every column of each region divided by the sought surfaces.

The whole network loss $L = L_{GDice}+L_{Div} +L_{smooth}+wL_{1} $, where $w=10$ is a weight coefficient for countering weak gradient when the prediction is close to the ground truth.

\begin{figure}
	\begin{subfigure}[b]{0.48\textwidth}
		\includegraphics[width=\textwidth]{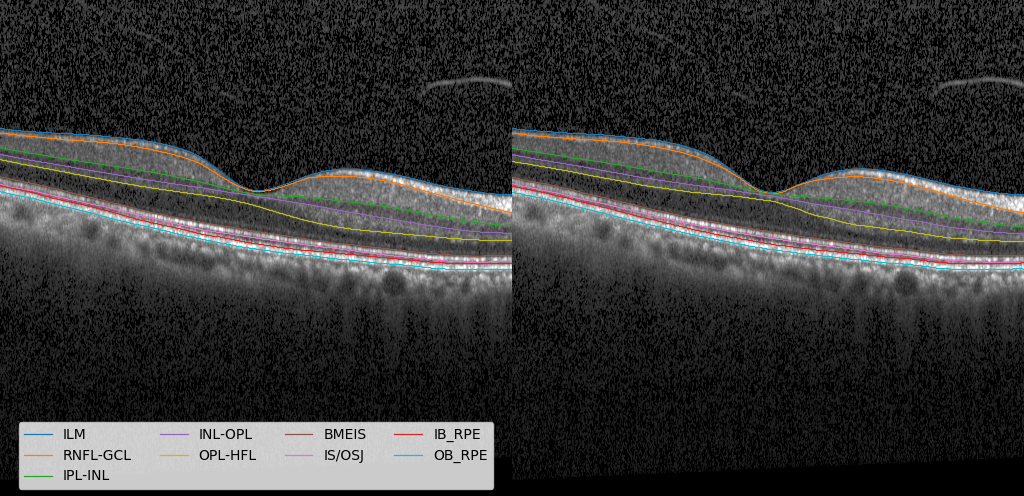}
		\caption{Segmentation of 9 intraretinal surfaces in an SD-OCT image of BES dataset.}
		\label{BESVisualResult}
	\end{subfigure}
	\hfil
	\begin{subfigure}[b]{0.48\textwidth}
		\includegraphics[width=\textwidth]{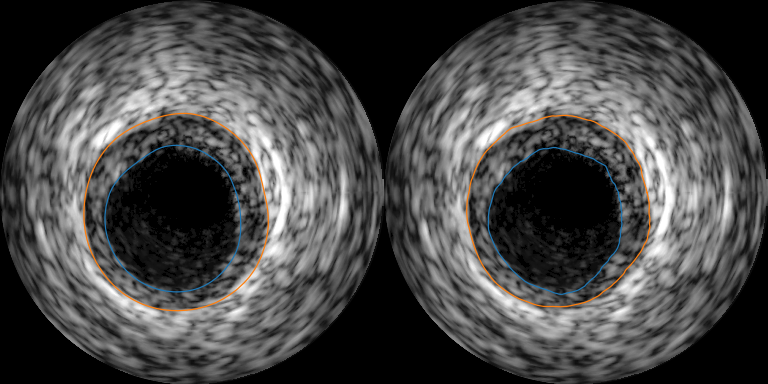}
		\caption{Segmentation results of lumen (blue) and media (orange) in an IVUS image. }
		\label{IVUSVisualResult}
	\end{subfigure}
	\caption{Sample segmentation on BES and IVUS dataset. In each subfigure, GT (L) and predictions (R).}
\end{figure}

\section{Experiments}
The proposed method was validated on two Spectral Domain Optical Coherence Tomography (SD-OCT) datasets for segmenting 9  retinal surfaces, and on one public Intravascular Ultrasound (IVUS) data set for the segmentation of lumen and media of vessel walls.

\subsection{SD-OCT Retinal Surface Segmentation}
 % experimenet name: expUnetTongren_20200313_9Surfaces_CV5_Sigma20
 \begin{table}[]
 	\caption{Mean Absolute Surface Distance (MASD) and standard deviation in $\mu m$ evaluated on Beijing Eye Study Dataset for segmenting 9 retinal surfaces. Below OE is OCT-Explorer\cite{OCTExplorer} graph search method, ours is the proposed method. Depth resolution is 3.87$\mu m$. }
 	% this OCTExplorer result is mean over whole data.
 	\label{BESExpResult}
 	\centering
 	\resizebox{\textwidth}{!}{%
 	\begin{tabular}{|c|c|c|c|c|c|c|c|c|c|c|}
 		\hline
 		&ILM&RNFL-GCL&IPL-INL&INL-OPL&OPL-HFL&BMEIS&IS/OSJ&IB\_RPE&OB\_RPE&Average\\
 		\hline
 		OE&1.79$\pm$4.34&3.58$\pm$4.75&2.92$\pm$4.77&2.54$\pm$4.77&2.73$\pm$4.72&1.79$\pm$4.74&8.61$\pm$5.35&1.82$\pm$4.72&1.78$\pm$4.72&3.06$\pm$5.15\\
 		%\hline
 		%FCBR-2\cite{YufanHe2019}&0.9745&2.8098&2.2552&2.2649&2.3933&1.2565&2.3985&1.3101&1.1489&1.8679&0.72637\\
 		ours  &0.98$\pm$0.09&2.98$\pm$0.41&2.59$\pm$0.47&2.38$\pm$0.43&2.70$\pm$0.65&1.43$\pm$0.49&2.82$\pm$0.70&1.53$\pm$0.28&1.21$\pm$0.19&2.07$\pm$0.91\\
 		\hline
 	\end{tabular}
 	}
 \end{table}
 
 \noindent
 {\bf Beijing Eye Study OCT Data set.} Beijing Eye Study 2011 has 3468 participants of aged 50+ years, but all of them have no segmentation ground truth. 47 health subjects without explicit eye diseases were randomly chosen, from which the graph-search based OCT-explorer 3.8 \cite{OCTExplorer} generated initial segmentation result, and then an experienced ophthalmologist manually corrected all 47 segmentation result as the ground truth for experiments. Choosing 47 subjects only is to save expensive ophthalmologist's correcting cost. All participants have scans on macula and optic nerve head by SD-OCT (Heidelberg Engineering, Inc., Germany) with a pixel resolution of 3.87~$\mu m$ in the height ($\vec{z}$-axis) direction. Each volume has scan composing of 31 single lines on the \ang{30}*\ang{30} field centered on the macula. Horizontal area of scan was reduced to \ang{20} centered on the macula to remove the optic disc region. 
\iffalse
\begin{wrapfigure}{l}{0.7\textwidth}
	%\centering
	%\includegraphics[width=\textwidth]{440_OD_5194_OCT16_Raw_GT_OCTExplorer_Predict.png}
	\includegraphics[width=0.7\textwidth]{120030_OD_3477_OCT15_GT_Predict.png}
	\caption{Simultaneous segmentation of 9 intraretinal surfaces in an SD-OCT image of BES data set. Ground truth (left) and predictions (right).}
	\label{BESVisualResult}
	%: 9 surfaces from top to bottom are ILM, RNFL-GCL, IPL-INL, INL-OPL, OPL-HFL, BMEIS, IS/OSJ, IB\_RPE, and OB\_RPE.
\end{wrapfigure}
\fi 
This experiment used a fixed $\sigma=20$ to generate the Gaussian ground truth, and used gaussian and pepper\&salt noises for data augmentation. A 10-fold cross-validation were performed to evaluate our method: 8 folds for training, 1 fold for validation, and 1 fold for testing. The mean absolute surface distances (MASDs) for each sought surface over the testing results on all 47 scans are shown in Table~\ref{BESExpResult}. Sample segmentation results are illustrated in Fig.~\ref{BESVisualResult}.

\iffalse
%current experiment result: expUnetJHU_20200310_sigma8_grad4_WeighedDiv10
\begin{wrapfigure}{l}{0.5\textwidth}
	%\centering
	\includegraphics[width=0.5\textwidth]{hc02_spectralis_macula_v1_s1_R_40_GT_Predict.png}
	%\includegraphics[width=\textwidth]{ms06_spectralis_macula_v1_s1_R_26_Image_GT_Predict.png}
	\caption{Simultaneous segmentation of 9 intraretinal surfaces in an SD-OCT image of JHU OCT dataset. Ground truth (left) and predictions (right). }
	\label{JHUVisualResult}
\end{wrapfigure}
\fi

\noindent
{\bf Public JHU OCT Dataset.} The public JHU retinal OCT data set \cite{JHUData2018} includes 35 human retina scans acquired on a Heidelberg Spectralis SD-OCT system, of which 14 are healthy controls (HC) and 21 have a diagnosis of multiple sclerosis (MS). Each patient has 49 B-scans with pixel size 496$\times$1024, and 9 ground truth surfaces on each B-Scan. The $\vec{z}$-axial resolution in each A-scan is 3.9~$\mu m$. The original images were manually delineated with 21 control points on each surface, and then a cubic interpolation was performed on each B-scan to obtain  the ground truth by a Matlab script~\cite{YufanHe2019}. Each B-scan was cropped to keep the  center 128 rows to from a 128$\times$1024 image.

The same data configuration and image input as in \cite{YufanHe2019} for training (6 HCs and 9 MS subjects) and testing (the remaining 20 subjects) were adopted in our experiment. A fixed $\sigma=8$ was used to generate Gaussian ground truth. Gaussian and pepper\&salt noises were used for data augmentation.  The MASDs for the proposed and He {\em et al.}'s methods are shown in Table~\ref{JHUExpResult}. While marginally improving the MASDs, our method demonstrates to be much more robust over the state-of-the-art He {\em et al.}'s method~\cite{YufanHe2019} with an improvement of 11.5\% on the standard deviation. Please refer to the supplementary material for the ablation experiments on this data set.
\begin{table}[]
	\caption{Average Absolute Surface error$\pm$StdDev (\SI{}{\micro\metre}) of JHU OCT Data. First 5 experiment results directly copy from \cite{YufanHe2019} table 1. The 6th experiment FCBR-2 is the re-implementation of paper \cite{YufanHe2019}. Bold font indicates the best in its row.}
	\label{JHUExpResult}
	\centering
	\resizebox{\textwidth}{!}{%
		\begin{tabular}{|l|l|l|l|l|l|l|l|}
			\hline
			Methods &AURA\cite{AURA2013}&R-Net\cite{R-NetOCT2018}&ReLayNet\cite{ReLayNet2017}&ShortPath\cite{YufanHe2019}&FCBR\cite{YufanHe2019}&FCBR-2\cite{YufanHe2019}&OurMethod\\
			\hline 
			ILM     &2.37$\pm$0.36 & 2.38$\pm$0.36 & 3.17$\pm$0.61 & 2.70$\pm$0.39 & 2.41$\pm$0.40&2.48$\pm$0.46 & \textbf{2.32}$\pm$ \textbf{0.27} \\
			RNFL-GCL&3.09$\pm$0.64 & 3.10$\pm$\textbf{0.55} & 3.75$\pm$0.84 & 3.38$\pm$0.68 & \textbf{2.96}$\pm$0.71&2.96$\pm$0.72& 3.07$\pm$0.68 \\
			IPL-INL &3.43$\pm$0.53 & 2.89$\pm$0.42 & 3.42$\pm$0.45 & 3.11$\pm$0.34 & 2.87$\pm$0.46&2.95$\pm$0.39& \textbf{2.86}$\pm$\textbf{0.33} \\
			INL-OPL &3.25$\pm$0.48 & 3.15$\pm$0.56 & 3.65$\pm$0.34 & 3.58$\pm$\textbf{0.32} & 3.19$\pm$0.53&\textbf{3.06}$\pm$0.45&3.24$\pm$0.60 \\ 
			OPL-ONL &2.96$\pm$0.55 & 2.76$\pm$0.59 & 3.28$\pm$0.63 & 3.07$\pm$\textbf{0.53} & \textbf{2.72}$\pm$0.61&2.92$\pm$0.73&2.73$\pm$0.57 \\
			ELM     &2.69$\pm$0.44 & 2.65$\pm$0.66 & 3.04$\pm$0.43 & 2.86$\pm$\textbf{0.41} & 2.65$\pm$0.73&\textbf{2.58}$\pm$0.55&2.63$\pm$0.51 \\
			IS-OS   &2.07$\pm$0.81 & 2.10$\pm$0.75 & 2.73$\pm$0.45 & 2.45$\pm$\textbf{0.31} & 2.01$\pm$0.57&\textbf{1.93}$\pm$0.75&1.97$\pm$0.57 \\
			OS-RPE  &3.77$\pm$0.94 & 3.81$\pm$1.17 & 4.22$\pm$1.48 & 4.10$\pm$1.42 & 3.55$\pm$1.02&\textbf{3.27}$\pm$\textbf{0.75}&3.35$\pm$0.83 \\
			BM      &2.89$\pm$2.18 & 3.71$\pm$2.27 & 3.09$\pm$\textbf{1.35} & 3.23$\pm$1.36 & 3.10$\pm$2.02&2.94$\pm$2.07&\textbf{2.88}$\pm$1.63 \\
			\hline
			Overall &2.95$\pm$1.04 & 2.95$\pm$1.10 & 3.37$\pm$0.92 & 3.16$\pm$0.88 & 2.83$\pm$0.99&2.79$\pm$0.96&\textbf{2.78}$\pm$\textbf{0.85} \\
			\hline
		\end{tabular}
	}
\end{table}

\subsection{IVUS Vessel Wall Segmentation}
\iffalse
\begin{wrapfigure}{l}{0.35\textwidth}
	%\centering
	\includegraphics[width=0.35\textwidth]{frame_05_0030_003_GT_Predict.png}
	%\includegraphics[width=\textwidth]{frame_10_0027_003_Image_GT_Predict.png}
	\caption{Segmentation results of lumen (blue) and media (orange) in an IVUS image. Ground truth (left) and predictions (right).}
	\label{IVUSVisualResult}
\end{wrapfigure}
\fi

\begin{table}[]
	\caption{Evaluation measurement $\pm$ stdDev of IVUS data. FCBR-2 is the re-implementation of paper\cite{YufanHe2019}. Bold indicates the best result in its comparison column. Blank cells mean un-reported result in original paper.}
	\label{IVUSExpResult}
	\centering
	\resizebox{\textwidth}{!}{%
		\begin{tabular}{|l|llll|llll|}
			\hline
			Methods      &            &\textbf{Lumen}          &       &                      &             &\textbf{Media}      &         &   \\
			\cline{2-9}          
			&Jacc        &Dice     &HD(\SI{}{\milli\metre})                    &PAD                   &Jacc         &Dice     &HD(\SI{}{\milli\metre})             &PAD  \\ 
			\hline
			GraphSearch\cite{GraphSearchIVUS2019}&0.86$\pm$0.04& &0.37$\pm$0.14 &0.09$\pm$0.03  &\textbf{0.90$\pm$0.03} & &0.43$\pm$0.12 &0.07$\pm$0.03 \\
			FCBR-2\cite{YufanHe2019}&\textbf{0.87$\pm$0.06} &\textbf{0.93$\pm$0.04}&0.43$\pm$0.37 &0.08$\pm$0.07 &0.89$\pm$0.07 &\textbf{0.94$\pm$0.04} &0.56$\pm$0.45 &0.07$\pm$0.07 \\
			OurMethod    &0.85$\pm$0.06&0.92$\pm$0.04&\textbf{0.36$\pm$0.20}&\textbf{0.08$\pm$0.06}&0.89$\pm$0.07&\textbf{0.94$\pm$0.04} &\textbf{0.40$\pm$0.30} &\textbf{0.06$\pm$0.06} \\
			\hline
		\end{tabular}
	}
\end{table}

The data used for this experiment was obtained from the standardized evaluation of IVUS image segmentation database~\cite{balocco2014standardized}. In this experiment, the data set B was used. This dataset consists  of 435 images with a size of $384\times384$, as well as the respective expert manual tracings  of lumen and  media surfaces. The pixel size is 0.026$\times$0.026. It comprises two groups - a training set (109 slices) and a testing set (326 slices). The experiment with the proposed method was conducted in conformance with the directives provided for the IVUS challenge. In our experiment, we randomly split the 109 training slices into 100 slices for training and 9 slices for validation. Each slice was transformed to be represented in the polar coordinate system with a size of $192\times 360$. Jaccard Measure (JM), Percentage of Area Difference (PAD) and Hausdroff Distance (HD) are utilized to evaluate segmentation accuracy, which are calculated using a Matlab script published in IVUS Challenge ~\cite{balocco2014standardized}. The results are summarized in Table~\ref{IVUSExpResult} comparing to the state-of-the-art automated methods. Sample segmentation results are illustrated in Fig.~\ref{IVUSVisualResult}. 

% our method is the experiment result: expUnetIVUS_Ablation_withLayerDice_20200304_1    

\section{Conclusion}
In this paper, a novel DL segmentation framework for multiple interacting surfaces is proposed with end-to-end training. The globally optimal solutions are achieved by seamlessly integrating two DL networks: one for surface cost parameterization with a Gaussian model and the other for total surface cost minimization while explicitly enforcing the surface mutual interaction constrains. The effectiveness of the proposed method was demonstrated on SD-OCT retinal layer segmentation and IVUS vessel wall segmentation. Though all our experiments were conducted on 2D, the method is ready for applications in 3D.

\bibliography{optimal_surface_seg_reference} 

\newpage
\appendix
\section{Ablation Experiments}
\begin{figure}
	\begin{subfigure}[b]{\textwidth}
		\includegraphics[width=\textwidth]{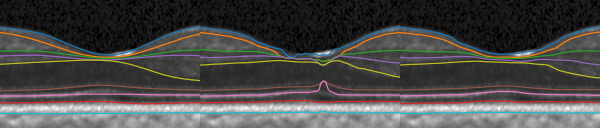}
		\caption{GT -- NoSmoothLoss -- WithSmoothLoss}
		\label{fig:SmoothLossAblation}
	\end{subfigure}
	%\hfil
	
	\begin{subfigure}[b]{\textwidth}
		\includegraphics[width=\textwidth]{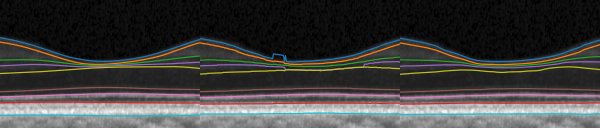}
		\caption{GT -- 1Channel -- 5Channels}
		\label{fig:IChannelAblation}
	\end{subfigure}
	
	\begin{subfigure}[b]{\textwidth}
		\includegraphics[width=\textwidth]{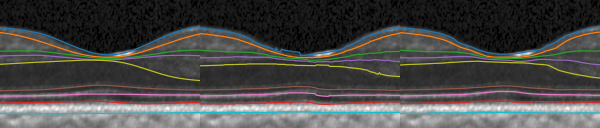}
		\caption{GT -- KLDivLoss -- WeightedDivLoss}
		\label{fig:KLDivAblation}
	\end{subfigure}
	
	\begin{subfigure}[b]{\textwidth}
		\includegraphics[width=\textwidth]{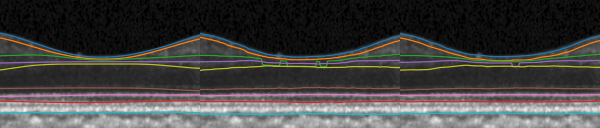}
		\caption{GT -- NoIPM -- WithIPM}
		\label{fig:IPMAblation}
	\end{subfigure}
	
	\caption{Sample segmentations on JHU OCT data for ablation 1 in (a), ablation 2 in (b), ablation 6 in (c), and ablation 7 in (d). All images are cropped A-Scan 400 to 600 in JHU OCT test set.}
\end{figure}

We did 7 ablation experiments on public JHU OCT data to verify our architecture design choices. Ablation experiment results are reported at table~\ref{AblationExpResult}. Ablation experiment 1 shows SmoothLoss improved both mean error and variance, and Fig.~\ref{fig:SmoothLossAblation} shows the explicit visual difference between with and without SmoothLoss. Ablation experiment 2 shows that adding gradient channels in input improved both mean error and variance, and Fig.~\ref{fig:IChannelAblation} is its visual result. Ablation 3 shows the value of dynamic sigma, which increases  error variance, but does not hurt mean performance; dynamic sigma may reduce the training effort of finding a proper fixed $\sigma$. Ablation 4 shows weighted surface cross entropy is  a little better than paper ~\cite{YufanHe2019} of general cross entropy, but not better than our new suggested weighted divergence Loss. Ablation 5 shows that the referred surface $\vec{\mu}$ from RegionConv will reduce error variance at 13\%, which makes sense that when network training enters stable stage, its region segmentation information is very helpful to prevent big surface errors. Ablation experiment 6 shows that general KLDiv Loss is worse than weighted divergence loss as weighted divergence loss more cares explicit gradient change points; a visual example is at Fig~\ref{fig:KLDivAblation}. Ablation experiment 7 and its visual Fig.~\ref{fig:IPMAblation} shows IPM model can improve surface dislocations. The prediction result of ablation experiment 7 without IPM optimization module has 361 pixels distributed in 10 slices of test set of violating separation constraints, which shows that network without IPM module can not learn implicit constraints in the ground truth, and IPM module has the capability of conforming constraints, reducing error, and reducing standard deviation (20\%).

\begin{table}[] 
	\label{AblationExpResult}	
	\caption{Average Absolute Surface error$\pm$StdDev (\SI{}{\micro\metre}) of ablation experiments on JHU OCT Data. Ablation-1 didn't not use SmoothLoss; Ablation-2 used single input channel, instead of 5 channels. Ablation-3 used dynamic $\vec{\sigma}$, instead of fixed $\vec{\sigma}=8$. Ablation-4 used weighted surface cross entropy, instead of weighted divergence loss. Ablation-5 didn't use the cooperation of referring surface $\vec{\mu}$ from region segmentation. Ablation-6 used KLDivLoss instead of weighted DivLoss. Ablation-7 didn't use IPM module.}
	\centering
	\tiny 
	\begin{tabular}{|l|l|l|l|l|l|l|l|l|}
		\hline
		Methods &Ablation-1&Ablation-2&Ablation-3&Ablation-4&Ablation-5&Ablation-6&Ablation-7&OurMethod\\
		\hline 
		ILM     &2.44$\pm$0.38 & 2.42$\pm$0.36 & 2.34$\pm$0.29 & 2.36$\pm$0.31 &2.42$\pm$0.32 &2.50$\pm$0.48 &2.36$\pm$0.38 &2.32$\pm$0.27 \\
		RNFL-GCL&3.12$\pm$0.64 & 3.25$\pm$0.84 & 2.97$\pm$0.65 & 2.98$\pm$0.69 &2.95$\pm$0.67 &3.24$\pm$0.89 &2.89$\pm$0.66 &3.07$\pm$0.68 \\
		IPL-INL &3.00$\pm$0.46 & 3.03$\pm$0.43 & 2.84$\pm$0.34 & 2.90$\pm$0.41 &2.88$\pm$0.56 &3.02$\pm$0.43 &2.86$\pm$0.44 &2.86$\pm$0.33 \\
		INL-OPL &3.25$\pm$0.58 & 3.27$\pm$0.43 & 3.05$\pm$0.42 & 3.21$\pm$0.61 &3.12$\pm$0.46 &3.25$\pm$0.49 &3.07$\pm$0.45 &3.24$\pm$0.60 \\ 
		OPL-ONL &2.82$\pm$0.53 & 2.88$\pm$0.71 & 2.77$\pm$0.60 & 2.71$\pm$0.57 &2.74$\pm$0.61 &3.26$\pm$0.80 &2.78$\pm$0.61 &2.73$\pm$0.57 \\
		ELM     &2.64$\pm$0.81 & 2.67$\pm$0.58 & 2.56$\pm$0.63 & 2.63$\pm$0.53 &2.66$\pm$0.45 &2.62$\pm$0.74 &2.67$\pm$0.61 &2.63$\pm$0.51 \\
		IS-OS   &2.01$\pm$0.81 & 1.95$\pm$0.71 & 2.05$\pm$0.78 & 1.98$\pm$0.71 &1.99$\pm$0.89 &2.02$\pm$0.61 &2.08$\pm$1.00 &1.97$\pm$0.57 \\
		OS-RPE  &3.40$\pm$0.71 & 3.45$\pm$0.83 & 3.35$\pm$0.81 & 3.51$\pm$0.85 &3.42$\pm$0.91 &3.64$\pm$0.82 &3.62$\pm$0.98 &3.35$\pm$0.83 \\
		BM      &2.83$\pm$1.80 & 2.90$\pm$1.88 & 3.07$\pm$2.10 & 2.89$\pm$2.05 &2.91$\pm$2.08 &2.96$\pm$1.32 &3.12$\pm$2.28 &2.88$\pm$1.63 \\
		\hline
		Overall &2.83$\pm$0.92 & 2.87$\pm$0.96 & 2.78$\pm$0.95 & 2.80$\pm$0.97 &2.79$\pm$0.98 &2.94$\pm$0.89 &2.83$\pm$1.06 &2.78$\pm$0.85\\
		\hline		
	\end{tabular}
\end{table}
% ablation-1 experiment name: expUnetJHU_20200313_sigma8_WeighedDiv10_NoSmoothLoss
% ablation-2 experiment name: expUnetJHU_20200309_Layer2SurfaceMu_sigma8
% ablation-3 experiment name: expUnetJHU_20200310_sigma0_grad4_WeighedDiv10
% ablation-4 experiment name: expUnetJHU_20200312_WeighedSurfaceCE_WeightedLayerCE
% ablation-5 expUnetJHU_20200316_sigma8_WeighedDiv10_NoReferringSurfFromLayer
% ablation-6 expUnetJHU_20200316_sigma8_KLDiv_InsteadWeighedDiv10
% ablation-7 experiment name: expUnetJHU_20200317_sigma8_WeighedDiv10_NoIPM
% our Method experiment name: expUnetJHU_20200310_sigma8_grad4_WeighedDiv10

\section{Sample Segmentations on 3 data set}
\begin{figure}
	\centering
	\includegraphics[width=\textwidth]{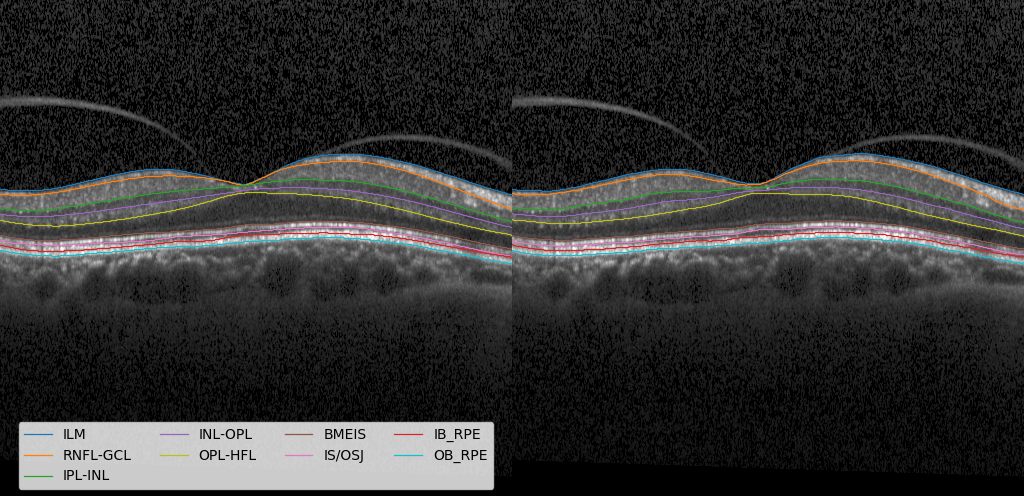}
	\caption{Simultaneous segmentation of 9 intraretinal surfaces in an SD-OCT image of BES OCT dataset. Ground truth (left) and predictions (right). }
	\label{fig:BESVisual2}
\end{figure}

\begin{figure}
	\centering
	\includegraphics[width=\textwidth]{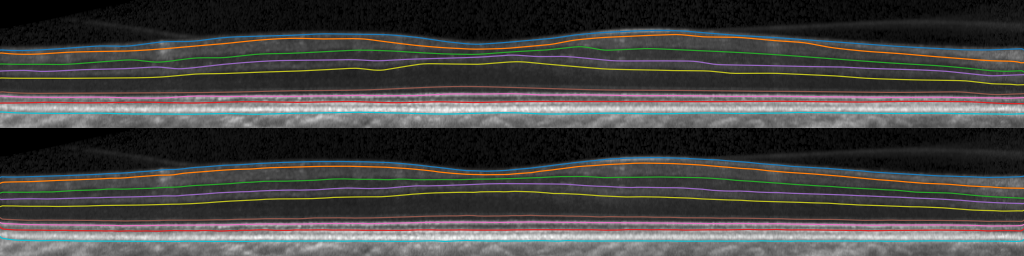}
	\caption{Simultaneous segmentation of 9 intraretinal surfaces in an SD-OCT image of HJU OCT dataset. Ground truth (top) and predictions (bottom). }
	\label{fig:HJUVisual2}
\end{figure}

\begin{figure} 
	\begin{subfigure}[b]{\textwidth}
		\includegraphics[width=\textwidth]{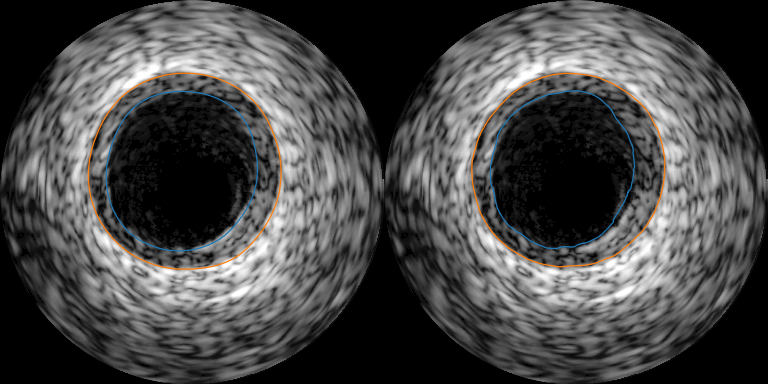}
		\caption{Good Sample}
		\label{fig:IVUSVisual2_good}
	\end{subfigure}
	%\hfil
	
	\begin{subfigure}[b]{\textwidth}
		\includegraphics[width=\textwidth]{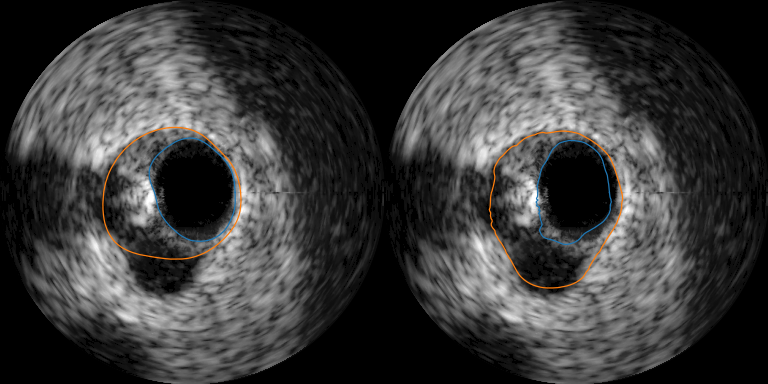}
		\caption{Bas Sample}
		\label{fig:IVUSVisual2_bad}
	\end{subfigure}
	
	\caption{Sample segmentations of lumen (blue) and media (orange) on IVUS data. Ground truth(left) and prediction(right).}
\end{figure}

Fig.~\ref{fig:BESVisual2}, ~\ref{fig:HJUVisual2}, ~\ref{fig:IVUSVisual2_good},  ~\ref{fig:IVUSVisual2_bad} gave some sample segmentations.

\section{IPM Optimization}
The 3D theoretical model in the previous problem formulation section needs to convert into a column model to implement. In this 3D theoretical model, the $\vec{S}$, $\vec{\mu}$, and $\vec{\sigma}$ all have a same size of $R^{N \times X \times Y}$, a 3D tensor. The implementation of this proposed segmentation framework in 3D is computationally demanding and memory hungry. In our experiments, we implemented it with 2D input slice. Multi-surfaces in a 2-D image can be viewed a series of smooth curves. A feasible surface intersects with each image column ($\mathbf{z}$-axis in 3D, or H dimension in 2D) exactly once. The space between 2 adjacent surfaces and boundary is defined as region, and surfaces should not cross each other. No surface should cross each other is, in mathematics, $\vec{s}_i \le \vec{s}_{i+1},\, \forall i\in[0,N-1)$ where $i$ indicating different surface index along the H dimension, and N is the number surface, $\vec{s}$ is the optimized surfaces location. An implemented math model is to build an optimization model on each column $q(x, y)$ along $\mathbf{z}$-axis, and then to parallelize this column model on GPU for all columns $q(x, y)$ to achieve the original 3D model. 

In our following notation, bold characters or numbers indicate vector or matrix, Diag means diagonal matrix. $\vec{s}= [s_0, s_1,..., s_{N-1}]^T$ expresses N surfaces along a column $q(x, y)$, similar convention also for $\vec{\mu}$, $\vec{\sigma}$, etc.

$S_i(q) - S_j(q)$ in previous problem formulation section expresses gap width between any 2 surfaces. It is enough in implementation to consider a simpler constraint of gap width between any 2 adjacent surfaces, like the A in formula ~\ref{eq:A}, which reduces the number of constraints from $N(N-1)/2$ to $(N-1)$.

\begin{equation} \label{eq:A}   
\bm{A} = 
\begin{bmatrix}
1  &-1  	&0     	  &\cdots &0 \\ 
0  &1   	&-1       &\cdots &0 \\ 
&    	&\ddots &\ddots   &  \\ 
0  &\cdots  &0  &1      &-1       \\ 
\end{bmatrix},
\end{equation}
where A is a constant matrix with dimension $(N-1)*N$. We also define a matrix $\bm{B} = \left[ \begin{smallmatrix} A \\ -A \end{smallmatrix} \right]$, where B is also a constant matrix with dimension of  $2(N-1)\times N$.

In general image format of implementation, $\mathbf{z}$-axis points downward, so a constraint condition can express as $\vec{b}_2 \le A\vec{s} \le \vec{b}_1$, where both $\vec{b}_1$ and $\vec{b}_2$ are a non-positive value vector of size $(N-1)$. $\vec{b}_2 \le A\vec{s} \le \vec{b}_1$ can further express as $\bm{B}\vec{s} = \left[\begin{smallmatrix} A \\ -A \end{smallmatrix} \right]\vec{s} \le \left[\begin{smallmatrix} \vec{b}_1 \\ -\vec{b}_2 \end{smallmatrix} \right]  = \bm{B}$, where $\bm{B}$ is a vector of size $2(N-1)$.

Now we define our implemented column model.  We expect final surfaces location $\vec{s}$ satisfying constraints, and has minimal deviation sum from initial prediction $\vec{\mu}$ with confidence index $\vec{\sigma}^2$ which are come from image feature analysis. In other words, under the constraint of  $\vec{s}_i \le \vec{s}_{i+1},\, \forall i\in[0,N-1)$, $\vec{s}$ may deviate a little bigger from $\vec{\mu}$ when $\vec{\sigma}^2$ is big; while $\vec{s}$ can deviate a little smaller from $\vec{\mu}$ when $\vec{\sigma}^2$ is small. Therefore, an matrix form constrained optimization model (column model) is constructed like below:

\begin{subequations} \label{eq:model_column}
\begin{align}
\vec{s}^* = &\operatorname*{argmin}_{\vec{s}}\frac{1}{2}(\vec{s}-\vec{\mu})^T\bm{Q}(\vec{s}-\vec{\mu}), \label{eq:model_column_cost}\\
&subject\,to\quad \bm{A}\vec{s} \le \vec{0}, \label{eq:model_column_A} \\
&\textbf{or}\,subject\,to\quad \bm{B}\vec{s} \le \bm{B}, \label{eq:model_column_B} 
\end{align}
\end{subequations}
where $\vec{s}= [s_0, s_1,..., s_{N-1}]^T$, $\vec{\mu}= [\mu_0, \mu_1,..., \mu_{N-1}]^T$, $\bm{Q}=Diag[\frac{1}{\sigma_0^2}, \frac{1}{\sigma_1^2},...,\frac{1}{\sigma_{N-1}^2}]$, $\bm{B}= \left[\begin{smallmatrix} \vec{b}_1 \\ -\vec{b}_2 \end{smallmatrix} \right]$ expresses the region gap range along a column, and $\vec{s}^*$ expresses final optimized solution. 

$\bm{A}\vec{s} \le \vec{0}$ is a special constraint case of more general $\bm{B}\vec{s} \le \bm{B}$ case. Following subsections A.1 and A.2 first consider above simple constrained case $\bm{A}\vec{s} \le \vec{0}$, and subsection A.3 considers more general constraint case $\bm{B}\vec{s} \le \bm{B}$.

\subsection{IPM Forward Optimization for $\bm{A}\vec{s} \le \vec{0}$}
\iffalse
Therefore, the system architecture for solving this surface segmentation can constructed as 3 parts: getting learning image features, reference of  possible $\vec{\mu}$ and $\vec{\sigma}^2$ from image features, then solving above constrained optimization model equation~\ref{eq:model_column}. And we hope these 3 parts may end-to-end(global) learn and optimize.
\fi
In order to solve above constrained convex optimization problem formula~\ref{eq:model_column_cost} with constraint $\bm{A}\vec{s} \le \vec{0}$, we write it into a Lagrangian form:
\begin{equation} \label{eq:Lagrangian}   
L(\vec{s},\vec{\lambda})= \frac{1}{2}(\vec{s}-\vec{\mu})^T\bm{Q}(\vec{s}-\vec{\mu})+\vec{\lambda}^T\bm{A}\vec{s},
\end{equation}
where $\vec{\lambda} \in R^{N-1}$.

Its corresponding perturbed KKT conditions are like below: 
\begin{equation} \label{eq:KKT_stationary}   
\text{Stationary: } \quad \bm{Q}(\vec{s}^*-\vec{\mu}) +\bm{A}^T\vec{\lambda}^* = \vec{0}
\end{equation}
\begin{equation}  \label{eq:KKT_slackness}  
\text{Perturbed Complementary Slackness: } \quad -Diag(\vec{\lambda}^*)\bm{A}\vec{s}^* = \frac{\vec{1}}{t}
\end{equation}
\begin{equation}  \label{eq:KKT_primal}  
\text{Primal feasibility: } \quad \bm{A}\vec{s}^* \le \vec{0}
\end{equation}
\begin{equation}  \label{eq:KKT_dual}  
\text{Dual feasibility: } \quad \vec{\lambda}^* \ge \vec{0},
\end{equation}
where $t$ is a perturbed slackness scalar and $t > 0$, and $\vec{1} \in R^{N-1}$ in equation~\ref{eq:KKT_slackness}. Bigger $t$ means smaller dual gap between original model function~\ref{eq:model_column} and the Lagrangian formula ~\ref{eq:Lagrangian} $L(\vec{s},\vec{\lambda})$; and $\vec{s}^*$ and $\vec{\lambda}^*$ indicate the optimal solution for the Lagrangian $L(\vec{s},\vec{\lambda})$. Equations~\ref{eq:KKT_stationary} and ~\ref{eq:KKT_slackness} give very important relations between $\vec{s}^*$ and $\bm{Q}$, and between  $\vec{s}^*$ and $\vec{\mu}$, which can be utilized in the back-propagation of the big deep learning optimization when $\vec{s}^*$ and $\vec{\lambda}^*$ are optimal solutions. 

We further construct a residual equation of equation ~\ref{eq:KKT_stationary} and ~\ref{eq:KKT_slackness}:
\begin{equation}   \label{eq:r_t} 
\vec{r}_t(\vec{s},\vec{\lambda}) = 
\begin{bmatrix}
\bm{Q}(\vec{s}-\vec{\mu}) +\bm{A}^T\vec{\lambda} \\ 
-Diag(\vec{\lambda})\bm{A}\vec{s} - \frac{\vec{1}}{t}
\end{bmatrix}
\end{equation}
where $\vec{r}_t(\vec{s},\vec{\lambda}) \in R^{2N-1}$. When $\vec{r}_t(\vec{s},\vec{\lambda}) = \vec{0}$, $\vec{s}$ and $\vec{\lambda}$ get their optimal solutions $\vec{s} = \vec{s}^*$, $\vec{\lambda} = \vec{\lambda}^*$. 

\begin{algorithm}[H] \label{alg:IPMForward}
	\SetAlgoLined
	\SetKwInOut {Input} {Input}
	\SetKwInOut {Output} {Output}
	\Input{$\vec{\mu}$, $\vec{\lambda}>\vec{0}$, $\bm{Q}$, $\bm{A}$, $1>\beta_1>0$, $1>\beta_2>0$, $\beta_3>1$, $\epsilon>0$}
	\Output{$\vec{s}^*$, $\bm{J}^{-1}$}
	\BlankLine	
	$\vec{s}= LIS(\vec{\mu})$\;
	$N = length(\vec{s})$\;
	\While{True}{
		$\vec{s}_0 = \vec{s}$\;
		$\vec{\lambda}_0 = \vec{\lambda}$\;
		$t= -\frac{\beta_3(N-1)}{(\bm{A}\vec{s})^T\vec{\lambda}}$\; 
		$\vec{r}_{0t}(\vec{s},\vec{\lambda}) = \left[ \begin{smallmatrix}  	\bm{Q}(\vec{s}-\vec{\mu}) +\bm{A}^T\vec{\lambda} \\ 
		-Diag(\vec{\lambda})\bm{A}\vec{s} - \frac{\vec{1}}{t}  \end{smallmatrix} \right] $\;
		$\bm{J} = \left[ \begin{smallmatrix} \bm{Q}    &\bm{A}^T \\ 
		-Diag(\vec{\lambda})\bm{A}  &-Diag(\bm{A}\vec{s})  \end{smallmatrix} \right]$\;
		$\left[ \begin{smallmatrix} \triangle\vec{s}\\ \triangle\vec{\lambda} \end{smallmatrix} \right] = -\bm{J}^{-1}\vec{r}_{0t}$\;
		$\alpha = min(1,min({-\lambda_i}/\triangle\lambda_i | \triangle\lambda_i <0))$\;
		$\vec{s} = \vec{s}_0+\alpha\triangle\vec{s}$\;
		\While {$\bm{A}\vec{s} >\vec{0}$}{
			$\alpha  = \alpha \beta_1$\;
			$\vec{s} = \vec{s}_0+\alpha\triangle\vec{s}$\;
		}
		
		\While {$\|\vec{r}_t(\vec{s},\vec{\lambda})\| > (1-\beta_2\alpha)\|\vec{r}_{0t}\|$}{
			$\alpha  = \alpha \beta_1$\;
			$\vec{s} = \vec{s}_0+\alpha\triangle\vec{s}$\;
			$\vec{\lambda} = \vec{\lambda}_0+\alpha\triangle\vec{\lambda}$\;			
		}
		\textbf{if} $\|\vec{r}_t\|< \epsilon$, \textbf{then}  break;	
	}
	\textbf{return} $\vec{s}$ and $\bm{J}^{-1}$
	\caption{IPM Forward Propagation}
\end{algorithm}

Using Newton iteration method to find the root of $\vec{r}_t(\vec{s},\vec{\lambda}) = \vec{0}$, we can get the iterative optimization formula in the small IPM forward optimization like below:
\begin{equation}  \label{eq:newtonIteration}   
\begin{bmatrix}
\vec{s}\\ 
\vec{\lambda}
\end{bmatrix} \leftarrow
\begin{bmatrix}
\vec{s}\\ 
\vec{\lambda} 
\end{bmatrix} - \alpha
\begin{bmatrix}
\bm{Q}   &\bm{A}^T \\ 
-Diag(\vec{\lambda})\bm{A}  &-Diag(\bm{A}\vec{s})
\end{bmatrix}^{-1} \vec{r}_t(\vec{s},\vec{\lambda}),
\end{equation}
where $\alpha >0 $ is an iterative step length. And let
\begin{equation}  \label{eq:J}  
\bm{J} = 
\begin{bmatrix}
\bm{Q}   &\bm{A}^T \\ 
-Diag(\vec{\lambda})\bm{A}  &-Diag(\bm{A}\vec{s})
\end{bmatrix},
\end{equation}
where $\bm{J} \in R^{(2N-1)\times(2N-1)}$ is a Jacobian matrix of $\vec{r}_t(\vec{s},\vec{\lambda})$ with respect to $[\vec{s}, \vec{\lambda}]$. After IPM forward iteration ends, $\bm{J}^{-1}$ will save for reuse in backward propagation of big deep learning optimization, which saves expensive inverse computation of matrix of a size of $R^{(2N-1)\times(2N-1)}$.

Therefore, the IPM iterative formula~\ref{eq:newtonIteration} may further express as:
\begin{equation}  \label{eq:newtonIteration_2}  
\begin{bmatrix}
\vec{s}\\ 
\vec{\lambda}
\end{bmatrix} \leftarrow
\begin{bmatrix}
\vec{s}\\ 
\vec{\lambda} 
\end{bmatrix} - \alpha\bm{J}^{-1} \vec{r}_t(\vec{s},\vec{\lambda}) = 
\begin{bmatrix}
\vec{s}\\ 
\vec{\lambda} 
\end{bmatrix} +\alpha
\begin{bmatrix}
\triangle\vec{s}\\ 
\triangle\vec{\lambda} 
\end{bmatrix},
\end{equation}
where $\triangle\vec{s}$ and $\triangle\vec{s}$ express the improving direction of $\vec{s}$ and $\vec{\lambda}$, and  $\left[ \begin{smallmatrix} \triangle\vec{s}\\ \triangle\vec{\lambda} \end{smallmatrix} \right]=  -\bm{J}^{-1}\vec{r}_t(\vec{s},\vec{\lambda})$. 

Newton iterative method guarantees the stationary of the solution, but it does not guarantee the feasibility of solution. The another core idea of IPM is that finding optimal solution starts from an interior feasible point,  uses Newton method find iterative improving direction, and then uses linear search to find a proper step to make sure a new iterative point is still in the feasible domain and at same time reduces the norm of residual $\vec{r}_t(\vec{s},\vec{\lambda})$. Therefore, in this each step of linear search process, algorithm needs to make sure equations ~\ref{eq:KKT_primal} and ~\ref{eq:KKT_dual} hold. The detailed algorithm of forward IPM Iteration is illustrated in Algorithm~\ref{alg:IPMForward}. 

As the Newton iterative method requires an initial point nearby its final goal root, in Algorithm~\ref{alg:IPMForward} a parallel $LIS$ (Largest Increasing Sub-sequence algorithm) is used to find most matching initial surface locations from the initial prediction $\vec{\mu}$, and then fills the non-largest increasing sub-sequence points with its neighbor value to make an initial $\vec{s}$ is feasible. As the dual gap between original cost function and Lagrangian is less than $\frac{N-1}{t}$, and $t = -(\bm{A}\vec{s})^T\vec{\lambda}$ deduced from the perturbed complementary slackness equation~\ref{eq:KKT_slackness},  this algorithm gradually enlarges t by using 	$t= -\frac{\beta_3(N-1)}{(\bm{A}\vec{s})^T\vec{\lambda}}, \beta_3 > 1$ to reduce dual gap, in order  to  get more accurate optimal solution to original cost function.  In order to avoid $\vec{\lambda} = \vec{\lambda} + \alpha\triangle\vec{\lambda} < 0 $ when $\triangle\lambda_i < 0$, choose $\alpha = min(1,min({-\lambda_i}/\triangle\lambda_i | \triangle\lambda_i <0))$ to make sure $\vec{1} \ge \vec{\lambda} \ge \vec{0}$.            

\subsection{IPM Backward Propagation for $\bm{A}\vec{s} \le \vec{0}$}
When IPM forward iteration converges in the small IPM optimization process, $\vec{r}_t(\vec{s}^*,\vec{\lambda}^*) = \vec{0}$. Its matrix form is
\begin{equation}  \label{eq:rt_optimal}  
\vec{r}_t(\vec{s}^*,\vec{\lambda}^*) = 
\begin{bmatrix}
\bm{Q_c}(\vec{s}^*-\vec{\mu}) +\bm{A_c}^T\vec{\lambda}^* \\ 
-Diag(\vec{\lambda}^*)\bm{A_c}\vec{s}^* - \frac{\vec{1}}{t}
\end{bmatrix}=\vec{0}.
\end{equation}
Using total differential with respect to variables $\bm{Q_c}$,  $\vec{s}$, $\vec{\lambda}$, and $\vec{\mu}$ gets:
\begin{equation}    
\begin{bmatrix}
\bm{Q_c}   &\bm{A_c}^T \\ 
-Diag(\vec{\lambda}^*)\bm{A_c}  &-Diag(\bm{A_c}\vec{s}^*)
\end{bmatrix}
\begin{bmatrix}
d\vec{s}^*\\ 
d\vec{\lambda}
\end{bmatrix}=
\begin{bmatrix}
-d\bm{Q_c}(\vec{s}^*-\vec{\mu})+\bm{Q_c}d\vec{\mu}\\ 
\vec{0}
\end{bmatrix}.
\end{equation}
Using formula~\ref{eq:J} $\bm{J}$ to replace above left-most matrix gets
\begin{equation} \label{eq:As<0_fullDifferential}    
\bm{J}
\begin{bmatrix}
d\vec{s}^*\\ 
d\vec{\lambda}
\end{bmatrix}=
\begin{bmatrix}
-d\bm{Q_c}(\vec{s}^*-\vec{\mu})+\bm{Q_c}d\vec{\mu}\\ 
\vec{0}
\end{bmatrix}.
\end{equation}

In back-propagation of the big deep learning optimization, the backward input to IPM optimization module is $\frac{dL}{d\vec{s}^*} \in R^N$, where L means loss in the deep learning network. Let's define 2 variables $\vec{d_s} \in R^N$ and $\vec{d_{\lambda}} \in R^{N-1}$ like below (Note: $\vec{d_s} \ne d\vec{s}^*$, and $\vec{d_{\lambda}} \ne d\vec{\lambda}$ ):
\begin{equation} \label{eq:d_sd_lambda}   
\begin{bmatrix}
\vec{d_s}\\ 
\vec{d_{\lambda}}
\end{bmatrix}= -\bm{J}^{-T}
\begin{bmatrix}
\frac{dL}{d\vec{s}^*}\\ 
\vec{0}
\end{bmatrix}.
\end{equation}
Transposing above equation gets
\begin{equation} \label{eq:dLds}    
\begin{bmatrix}
(\frac{dL}{d\vec{s}^*})^T, \vec{0}^T
\end{bmatrix}\bm{J}^{-1}=-
\begin{bmatrix}
\vec{d_s}^T, \vec{d_{\lambda}}^T
\end{bmatrix}
\end{equation}
This equation~\ref{eq:dLds} will be used in the following backward gradient computation.

\textbf{Compute $\frac{dL}{d\vec{\mu}}$: } in the right side of full differential equation~\ref{eq:As<0_fullDifferential}, keep $d\vec{\mu}$ unchanged, and set other differentials equal zeros. Related deductions are like below (where superscripts like $M \times N $ etc indicate dimension of a matrix,  subscripts like $i$ indicate the index of a component, $I$ are an identity matrix,  $M=N-1$ in this case, and following are same):
\begin{equation*}       
\begin{aligned}
\bm{J}
\begin{bmatrix}
d\vec{s}^*\\ 
d\vec{\lambda}^*
\end{bmatrix}
&=
\begin{bmatrix}
\bm{Q_c} d\vec{\mu}\\ 
\vec{0}^{M\times 1}
\end{bmatrix},\\ % end equation 1
\bm{J}
\begin{bmatrix}
\frac{d\vec{s}^*}{d\vec{\mu}}\\ 
\frac{d\vec{\lambda}^*}{d\vec{\mu}}
\end{bmatrix}
&=
\begin{bmatrix}
\bm{Q_c}^{N \times N}\\ 
0^{M \times N}
\end{bmatrix},\\ % end equation 2
\begin{bmatrix}
(\frac{dL}{d\vec{s}^*})^T, \vec{0}^T
\end{bmatrix}\bm{J}^{-1}
\bm{J}
\begin{bmatrix}
\frac{d\vec{s}^*}{d\vec{\mu}}\\ 
\frac{d\vec{\lambda}^*}{d\vec{\mu}}
\end{bmatrix}
&=-
\begin{bmatrix}
\vec{d_s}^T, \vec{d_{\lambda}}^T
\end{bmatrix}
\begin{bmatrix}
\bm{Q_c}^{N \times N}\\ 
0^{M \times N}
\end{bmatrix},\\ % end equation 3
(\frac{dL}{d\vec{s}^*})^T \frac{d\vec{s}^*}{d\vec{\mu}}
&=-\vec{d_s}^T \bm{Q_c}, \\%next equation 4
(\frac{dL}{d\vec{\mu}})^T
&=-\vec{d_s}^T \bm{Q_c}. \\ % end equation 5
\end{aligned}
\end{equation*}
It finally gets: 
\begin{equation} \label{eq:AS<0_dLdmu}
\frac{dL}{d\vec{\mu}}=-\bm{Q_c}^T \vec{d_s}.      
\end{equation}

\newpage 

\begin{algorithm}[H] \label{alg:IPMBackward}
	\SetAlgoLined
	\SetKwInOut {Input} {Input}
	\SetKwInOut {Output} {Output}
	\Input{$\frac{dL}{d\vec{s}^*}$, $\bm{J}^{-1}$, $\vec{s}^*$, $\bm{Q_c}$, $\vec{\mu}$,}
	\Output{$\frac{dL}{d\bm{Q_c}}$, $\frac{dL}{d\vec{\mu}}$}
	\BlankLine	
	
	$\left[ \begin{smallmatrix} \vec{d_s}\\ \vec{d_{\lambda}} \end{smallmatrix} \right] = -\bm{J}^{-T}
	\left[ \begin{smallmatrix}
	\frac{dL}{d\vec{s}^*}\\ \vec{0}
	\end{smallmatrix} \right]$\;
	$\frac{dL}{d\bm{Q_c}}=Diag(\vec{d_s}){\begin{bmatrix}
		(\vec{s}^*-\vec{\mu})^T\\
		(\vec{s}^*-\vec{\mu})^T\\
		\vdots\\
		(\vec{s}^*-\vec{\mu})^T\\
		\end{bmatrix}
	}^{N \times N}$\; 
	$\frac{dL}{d\vec{\mu}}= -\bm{Q_c}^T \vec{d_s} $\;
	\caption{IPM Backward Propagation Algorithm}
\end{algorithm}

\textbf{Compute $\frac{dL}{d\bm{Q_c}}$: } in the right side of full differential equation~\ref{eq:As<0_fullDifferential}, keep $d\bm{Q_c}$ unchanged, set other differentials equal zeros, and decompose $\bm{Q_c} \in R^{N \times N}$ into  
$N$ rows where each row is a vector $\vec{q}_i^T$, where $\vec{q}_i \in R^N$ and $i \in [0,N)$, that $\bm{Q_c} = [\vec{q}_0, \vec{q}_1, \vec{q}_2, \cdots, \vec{q}_{N-1}]^T$ like below. Related deductions are like below :
\begin{equation*}
\begin{aligned}   
\bm{J}
\begin{bmatrix}
d\vec{s}^*\\ 
d\vec{\lambda}^*
\end{bmatrix}
&=
\begin{bmatrix}
-d{\bm{Q_c}}(\vec{s}^*-\vec{\mu})\\ 
0^{M \times 1}
\end{bmatrix},\\% end equation 1
\bm{J}
\begin{bmatrix}
d\vec{s}^*\\ 
d\vec{\lambda}^*
\end{bmatrix}
&=
\begin{bmatrix}
-{\begin{bmatrix}
	d\vec{q}_0^T\\
	d\vec{q}_1^T\\
	d\vec{q}_2^T\\
	\vdots\\
	d\vec{q}_{N-1}^T\\	
	\end{bmatrix}
}(\vec{s}^*-\vec{\mu})\\
\\ 
0^{M \times 1}
\end{bmatrix}.\\% end equation 2
\end{aligned} 
\end{equation*}
In order to get partial differential with respect to $\vec{q}_i \in R^N$, let $\bm{Q}_j = \vec{0}$ where $j \ne i$ and $i,j \in [0,N)$, getting
\begin{equation*}
\begin{aligned}   
\bm{J}
\begin{bmatrix}
d\vec{s}^*\\ 
d\vec{\lambda}^*
\end{bmatrix}
&=
\begin{bmatrix}
-{\begin{bmatrix}
	\vec{0}^T\\
	\vec{0}^T\\
	\vdots\\
	d\vec{q}_{i}^T\\
	\vdots\\
	\vec{0}^T\\	
	\end{bmatrix}
}^{N \times N}(\vec{s}^*-\vec{\mu})\\
\\ 
0^{M \times 1}
\end{bmatrix},\\% end equation 1
\bm{J}
\begin{bmatrix}
d\vec{s}^*\\ 
d\vec{\lambda}^*
\end{bmatrix}
&=
\begin{bmatrix}
{\begin{bmatrix}
	0\\
	0\\
	\vdots\\
	-(\vec{s}^*-\vec{\mu})^T d\vec{q}_i\\
	\vdots\\
	0\\	
	\end{bmatrix}
}^{N \times 1}\\
\\ 
0^{M \times 1}
\end{bmatrix},\\% end equation 2
\bm{J}
\begin{bmatrix}
\frac{d\vec{s}^*}{d\vec{q}_i}\\ 
\frac{d\vec{\lambda}^*}{d\vec{q}_i}
\end{bmatrix}
&=
\begin{bmatrix}
{\begin{bmatrix}
	\vec{0}^T\\
	\vec{0}^T\\
	\vdots\\
	-(\vec{s}^*-\vec{\mu})^T \\
	\vdots\\
	\vec{0}^T\\
	\end{bmatrix}
}^{N \times N}\\
\\ 
0^{M \times N}
\end{bmatrix},\\% end equation 3
\begin{bmatrix}
(\frac{dL}{d\vec{s}^*})^T, \vec{0}^T
\end{bmatrix}\bm{J}^{-1}
\bm{J}
\begin{bmatrix}
\frac{d\vec{s}^*}{d\vec{q}_i}\\ 
\frac{d\vec{\lambda}^*}{d\vec{q}_i}
\end{bmatrix}
&=-
\begin{bmatrix}
\vec{d_s}^T, \vec{d_{\lambda}}^T
\end{bmatrix}
\begin{bmatrix}
{\begin{bmatrix}
	\vec{0}^T\\
	\vec{0}^T\\
	\vdots\\
	-(\vec{s}^*-\vec{\mu})^T \\
	\vdots\\
	\vec{0}^T\\
	\end{bmatrix}
}^{N \times N}\\
\\ 
0^{M \times N}
\end{bmatrix},\\% end equation 4
(\frac{dL}{d\vec{s}^*})^T \frac{d\vec{s}^*}{d\vec{q}_i}
&=(\vec{d_s})_i (\vec{s}^*-\vec{\mu})^T,\\% end equation 5
\frac{dL}{d\vec{q}_i^T} &= (\vec{d_s})_i (\vec{s}^*-\vec{\mu})^T. \\% end equation 6
\end{aligned} 
\end{equation*}
Now, combining all $\frac{dL}{d\vec{q}_i^T} \in R^{1 \times N}$ of $\bm{Q_c}$ gets
\begin{equation} \label{eq:AS<0_dLdQ}
\frac{dL}{d\bm{Q_c}}= Diag(\vec{d_s}){\begin{bmatrix}
	(\vec{s}^*-\vec{\mu})^T\\
	(\vec{s}^*-\vec{\mu})^T\\
	\vdots\\
	(\vec{s}^*-\vec{\mu})^T\\
	\end{bmatrix}
}^{N \times N}. 
\end{equation}  

Equations~\ref{eq:AS<0_dLdmu} and ~\ref{eq:AS<0_dLdQ} are exactly the back propagation loss gradient with respect to $\bm{Q_c}$ and $\vec{\mu}$ in the big deep learning optimization. Naturally, back propagation algorithm for IPM module in the big deep learning optimization is like algorithm ~\ref{alg:IPMBackward}:

We used Pytorch \cite{Pytorch2015} 1.3.1 implementing this IPM forward and backward propagation algorithm with batch-supported GPU parallelism in Ubuntu Linux. In our practice, we found at most 7 IPM iterations can achieve enough accuracy, with $\beta_1=0.5$, $\beta_2=0.055$, $\beta_3=10.0$, $\epsilon=0.01$. In our implementation, we use pseudo inverse to replace normal inverse in computing $\bm{J}^{-1}$ only when $\bm{J}$ is a singular matrix sometimes.

In order to facilitate further research and comparison on surface segmentation field, all our code including data pre-processing, core parallel IPM code, framework code, experiment configure(yaml) are public at \cite{MoDL-OSSeg}.  

\subsection{IPM Optimization for $\bm{B}\vec{s} \le \bm{B}$}
In order to solve above constrained convex optimization problem formula~\ref{eq:model_column_cost} with constraint $\bm{B}\vec{s} \le \bm{B}$, we write it into a Lagrangian form:
\begin{equation} \label{eq:Lagrangian_B}   
L(\vec{s},\vec{\lambda})= \frac{1}{2}(\vec{s}-\vec{\mu})^T\bm{Q}(\vec{s}-\vec{\mu})+\vec{\lambda}^T(\bm{B}\vec{s}-\bm{B}),
\end{equation}
where $\vec{\lambda} \in R^{2(N-1)}$, $\bm{B} \in R^{2(N-1)\times N}$, $\bm{B} \in R^{2(N-1)}$.

Its corresponding perturbed KKT conditions are like below: 
\begin{equation} \label{eq:KKT_stationary_B}   
\text{Stationary: } \quad \bm{Q}(\vec{s}^*-\vec{\mu}) +\bm{B}^T\vec{\lambda}^* = \vec{0}
\end{equation}
\begin{equation}  \label{eq:KKT_slackness_B}  
\text{Perturbed Complementary Slackness: } \quad -Diag(\vec{\lambda}^*)(\bm{B}\vec{s}^*-\bm{B}) = \frac{\vec{1}}{t}
\end{equation}
\begin{equation}  \label{eq:KKT_primal_B}  
\text{Primal feasibility: } \quad \bm{B}\vec{s}^* -\bm{B} \le \vec{0}
\end{equation}
\begin{equation}  \label{eq:KKT_dual_B}  
\text{Dual feasibility: } \quad \vec{\lambda}^* \ge \vec{0},
\end{equation}
where $t$ is a perturbed slackness scalar and $t > 0$, and $\vec{1} \in R^{2(N-1)}$ in equation~\ref{eq:KKT_slackness_B}. Bigger $t$ means smaller dual gap between original model function~\ref{eq:model_column} and the Lagrangian formula ~\ref{eq:Lagrangian_B} $L(\vec{s},\vec{\lambda})$; and $\vec{s}^*$ and $\vec{\lambda}^*$ indicate the optimal solution for the Lagrangian $L(\vec{s},\vec{\lambda})$. Equations~\ref{eq:KKT_stationary_B} and~\ref{eq:KKT_slackness_B} also give very important relations between $\vec{s}^*$ and $\bm{Q}$, between  $\vec{s}^*$ and $\vec{\mu}$, and between  $\vec{s}^*$ and $\bm{B}$, which can be utilized in the back-propagation of the big deep learning optimization when $\vec{s}^*$ and $\vec{\lambda}^*$ are optimal solutions. 

We further construct a residual equation of equation ~\ref{eq:KKT_stationary_B} and ~\ref{eq:KKT_slackness_B}:
\begin{equation}   \label{eq:r_t_B} 
\vec{r}_t(\vec{s},\vec{\lambda}) = 
\begin{bmatrix}
\bm{Q}(\vec{s}-\vec{\mu}) +\bm{B}^T\vec{\lambda} \\ 
-Diag(\vec{\lambda})(\bm{B}\vec{s}-\bm{B}) - \frac{\vec{1}}{t}
\end{bmatrix}
\end{equation}
where $\vec{r}_t(\vec{s},\vec{\lambda}) \in R^{3N-2}$. When $\vec{r}_t(\vec{s},\vec{\lambda}) = \vec{0}$, $\vec{s}$ and $\vec{\lambda}$ get their optimal solutions $\vec{s} = \vec{s}^*$, $\vec{\lambda} = \vec{\lambda}^*$. 

Following same deduction process in section A.1 and A.2, we can get IPM forward and backward formula. In application $\bm{B} = \left[\begin{smallmatrix} \vec{b}_1 \\ -\vec{b}_2 \end{smallmatrix} \right]$ is different in each column $q(x,y)$, and it is general a learning vector from previous layers in a network, so following formula deduction needs to compute $\frac{dL}{\vec{db}}$ with a similar deduction process of $\frac{dL}{\vec{d\mu}}$, and then maps back to  $\frac{dL}{\vec{db_1}}$ and $\frac{dL}{\vec{db_2}}$. Readers also can refer OptNet~\cite{OptNet2017} for its following deduction.

\end{document}